\documentclass{article}

\PassOptionsToPackage{numbers, compress}{natbib}
\usepackage[preprint]{neurips}

\usepackage[utf8]{inputenc} 
\usepackage[T1]{fontenc}    
\usepackage{hyperref}       
\usepackage{url}            
\usepackage{booktabs}       
\usepackage{amsfonts}       
\usepackage{nicefrac}       
\usepackage{microtype}      
\usepackage{xcolor}         
\usepackage{bbding}
\usepackage{wrapfig}
\usepackage{graphicx}
\usepackage{upquote}
\usepackage{subcaption}
\usepackage{multirow}

\usepackage{textcomp}  
\usepackage{scalerel}  
\usepackage{xspace}
\usepackage{amsmath}

\newcommand{\modelname}{MARIO\xspace}

\title{\modelname: MAth Reasoning with code Interpreter Output\\-- A Reproducible Pipeline}

\author{%
  Minpeng Liao\thanks{equal contribution}~, Wei Luo$^*$, Chengxi Li$^*$, Jing Wu$^*$, Kai Fan\thanks{corresponding author}\\
  Alibaba Group \\
  \texttt{\{minpeng.lmp,muzhuo.lw,xiji.lcx,wj334275,k.fan\}@alibaba-inc.com} \\
}

\begin{document}

\maketitle

\begin{abstract}

Large language models (LLMs) have seen considerable advancements in natural language understanding tasks, yet there remains a gap to bridge before attaining true artificial general intelligence, especially concerning shortcomings in mathematical reasoning capabilities. 
We postulate that the inherent nature of LLM training, which focuses on predicting probabilities of next token, presents challenges in effectively modeling mathematical reasoning that demands exact calculations, both from data-driven and theoretical standpoints.
In this paper, we address this challenge by enriching the data landscape and introducing a novel math dataset, enhanced with a capability to utilize a Python code interpreter. 
This dataset is derived from GSM8K\cite{gsm8kcobbe2021} and MATH\cite{mathhendrycks2021} and has been further refined through a combination of GPT-4 annotations, human review, and self-training processes, where the errors in the original GSM8K training set have been fixed. 
Additionally, we propose a tentative, easily replicable protocol for the fine-tuning of math-specific LLMs, which has led to a significant improvement in the performance of a 7B-parameter LLM on the GSM8K and MATH datasets. 
We are committed to advancing the field of mathematical reasoning in LLMs and, to that end, we have made source code for data generation / training / inference, and the model checkpoints publicly available at \url{https://github.com/MARIO-Math-Reasoning/MARIO}. 
We hope this will facilitate further research and development within the community.

\end{abstract}

\section{Introduction}
\label{sec:intro}

The Chain-of-Thought (CoT) prompting technique~\cite{wei2022chain} has been empirically shown to enhance the complex reasoning capabilities of large language models (LLMs) by generating a sequence of intermediate reasoning steps. 
Commercial or proprietary LLMs, including GPT-4\cite{openai2023gpt} and Claude-2~\cite{claude}, are designed to produce CoT responses by default, leading to improved mathematical reasoning performance, as evidenced by a 50.36\% accuracy rate on the MATH dataset with GPT-4~\cite{zhou2023solving}. 
Moreover, when these LLMs are augmented with a plugin capable of executing code snippets, their accuracy in arithmetic computations—which are typically challenging for LLMs—is further enhanced. 
For instance, GPT-4-Code achieved a 69.69\% accuracy rate on the MATH dataset~\cite{zhou2023solving}. 
This underscores the efficacy of integrating text analysis with code execution in datasets designed for mathematical reasoning tasks. 


\begin{figure}[t]
\centering
\begin{subfigure}{\textwidth}
    \includegraphics[width=\textwidth]{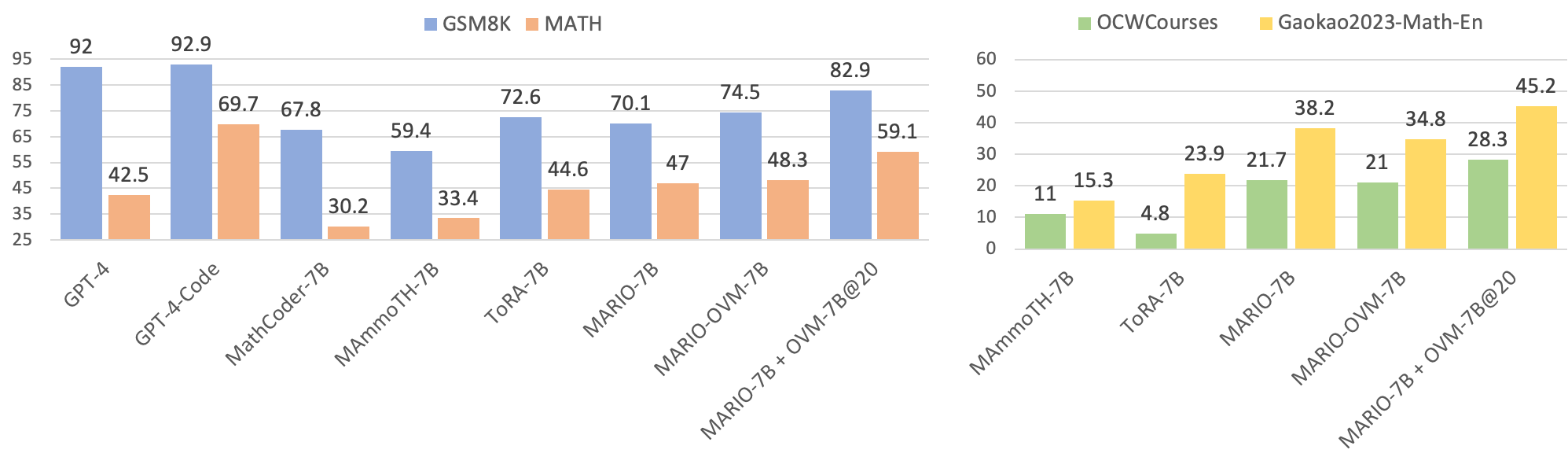}
    \caption{Zero-shot performance on in-domain (left) and out-of-domain (right) datasets. The results of GPT and ToRA on in-domain datasets are reported in \cite{zhou2023solving} and \cite{gou2023tora}, and the out-of-domain result is obtained by running experiments with its official released repository. For MARIO-7B + OVM-7B, 20 sampled solutions are generated from the MARIO-7B. For MARIO-OVM-7B, the OVM is used for a single solution generation.}
    \label{fig:intro_cmp}
\end{subfigure}
\hfill
\begin{subfigure}{\textwidth}
    \includegraphics[width=\textwidth]{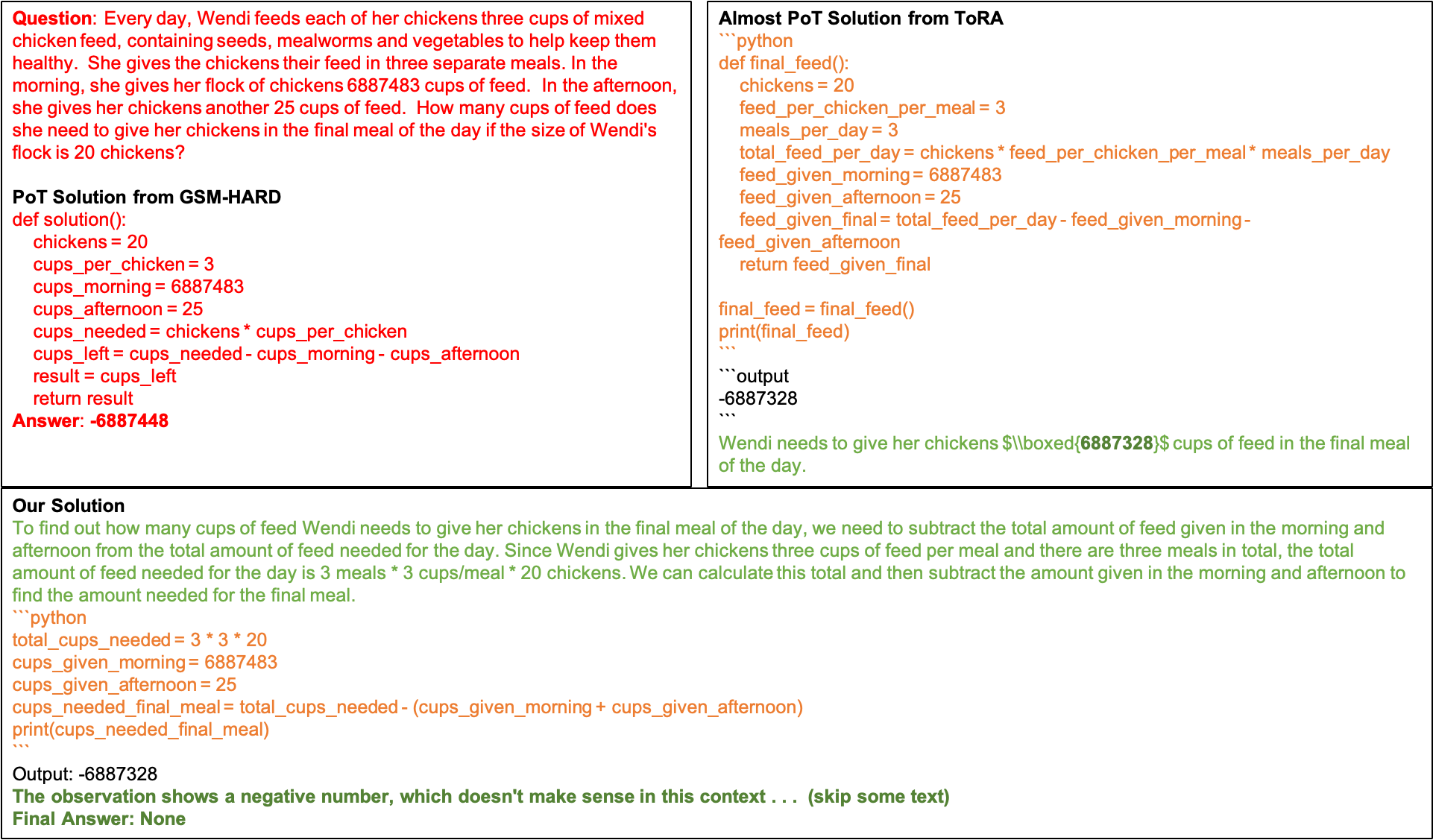}
    \caption{An example from an unreasonable question from GSM-Hard. In the PoT solution provided by GSM-Hard~\cite{gao2023pal}, the purely code-centric approach yields a negative quantity of cups. The almost PoT solution from ToRA omits the minus sign in its concluding sentence, failing to address the impractical issue. Our solution, incorporating both text analysis and a code snippet, recognizes that a negative quantity of cups is illogical.}
    \label{fig:intro_cmp1}
\end{subfigure}
\caption{Comparison of our approach with PoT style methods.}
\end{figure}

Recently, \citet{yang2023gpt} demonstrated that, despite extensive fine-tuning on arithmetic operations, large language models (LLMs) are still unable to achieve perfect accuracy. 
Consequently, integrating code for precise numerical computation has become an inevitable trend. 
The Program-of-Thought (PoT) framework~\cite{chen2022program} and Program-Aided Language (PAL) models~\cite{gao2023pal} represent seminal efforts in generating code-centric datasets for academic research. 
Building on this trend, ToRA~\cite{gou2023tora} advances the field by employing a proprietary annotation methodology along with GPT-4, setting a new benchmark for state-of-the-art performance in solving mathematical problems via code interpretation. 
However, it is noteworthy that the solutions contained within these datasets predominantly consist of code snippets, with minimal accompanying textual analysis. 
An obvious shortcoming of a code-centric solution is that it can overlook common sense in math word problems. 
For instance, as shown in Figure~\ref{fig:intro_cmp1}, the PoT solutions disregard the fact that the quantity of the food taken cannot be negative. 
MathCoder~\cite{wang2023mathcoder} represents an initiative that emulates the response patterns of GPT-4, integrating a plugin proficient in both code generation and natural language reasoning. 
This approach harnesses the capabilities of GPT-4, equipped with a code interpreter, to automate the annotation process, producing a blend of text analyses and code snippets, guided by appropriate instructions. 

Our data generation approach generally mirrors that of MathCoder, but with the enhancement of an extra layer of human verification specifically for the correction of easily rectifiable errors. 
Therefore, all erroneous solutions in the original GSM8K training set have been manually fixed. 
In contrast, the MATH dataset encompasses more challenging problems and necessitates the expertise of professional annotators. 
To address this, we utilize self-training and knowledge distillation techniques to selectively identify correct solutions, in conjunction with more samplings. 

Furthermore, we introduce a tentative and easily replicable protocol for the fine-tuning of math-specific Large Language Models (LLMs). 
To ensure the reproducibility of our experiments, we begin with the extensively studied LLM, Llama-2~\cite{touvron2023llama}, and utilize its math-oriented continual pre-training variant, Llemma~\cite{azerbayev2023llemma}. 
Subsequently, we apply supervised fine-tuning to our annotated dataset to establish a baseline model. 

Unlike text generation tasks such as summarization or translation, mathematical reasoning usually yields a unique answer, which simplifies the verification of its correctness. 
However, assessing the reasoning process that leads to the final answer remains challenging. 
To address this issue, \citet{lightman2023let} introduced a model supervised by processes and a corresponding dataset with manually labeled solution procedures. 
Yet, from a replication standpoint, creating these annotations is both labor-intensive and costly. 
As a compromise, we recommend training a straightforward outcome-supervised model with both correct and incorrect solutions as described by \cite{cobbe2021training,yu2023outcome}, to serve as an auxiliary tool for comparing and selecting the best among various solutions. 
Our outcome value model (OVM) is fine-tuned using the efficient LoRA training~\cite{hu2021lora} in a multi-task setting, which enables the model to conserve computational resources while maintaining its generative capabilities.

In summary, our main contributions are in three-fold.
\begin{enumerate}
    \item We create a math reasoning format that integrates both text analyses and code snippets, leveraging logical reasoning and precise computation.
    \item We introduce a reproducible pipeline for fine-tuning a large language model in the mathematical domain, which includes Continual Pre-training (CPT), Supervised Fine-tuning (SFT), and Multi-Task OVM Fine-tuning.
    \item Both our SFT and Value LLMs can generate solutions, and the value LLM can play the role of outcome value model. Our experiments demonstrate that our approach can significantly enhance performance on math reasoning tasks, as highlighted in Figure~\ref{fig:intro_cmp}. 
\end{enumerate}

\section{Dataset}
\label{sec:data}

In this section, we will provide a detailed description of our methodology and the steps involved in constructing the complete corpus. 
As mentioned earlier, the solution we aim to develop must integrate text analysis and code snippets seamlessly. 
The textual content should articulate the problem-solving process, while the code snippets should perform precise computations, which may encompass symbolic simplification, differentiation, numerical analysis, equation solving, and handling inequalities, among other tasks. 
Our starting point will be math problems at the grade school level, specifically the dataset GSM8K.

\textbf{GSM8K} To create a solution in the desired format, we utilize instructions inspired by REACT~\cite{yao2022react} to ensure GPT recognizes when to employ an external tool, such as an offline Python code interpreter, when needed. 
In addition to the REACT instruction in the prompt, we provide two manually crafted demonstration examples within the prompt for the language model to emulate. 
For further details on the prompt setup, please see Appendix~\ref{app:data}. 

Given that the problems in GSM8K are relatively straightforward, we initially prompt both GPT-3.5 and GPT-4 to tackle each question with at maximum 5 code snippets in the training set using a temperature 0, which is intended to simulate greedy decoding.\footnote{While we anticipate that a temperature setting of 0 would result in greedy decoding, it's worth noting that \texttt{gpt-3.5-turbo-0613} or \texttt{gpt-4-0613} with temperature 0 does not ensure deterministic outcomes.} 
For questions that remain unsolved after the initial two attempts, we address the potential requirement for more creative solutions by re-prompting GPT-4 with a temperature 0.6 for another two attempts. 
This approach is intended to enhance the diversity of the generated responses. 
Following this method, we successfully obtain at least one correct solution for 98.3\% of the questions in the GSM8K training set.

Approximately 100 questions remain for which the answers generated by GPT do not align with the answers from the original training set. 
Given the manageable number, we have conducted a manual review of the discrepancies and corrected any inaccuracies found in either the LLM-generated or the original answers. 
This ensures that each question within the GSM8K training set is associated with at least one correct solution in required format.

\begin{figure}[t]
    \centering
    \includegraphics[width=\textwidth]{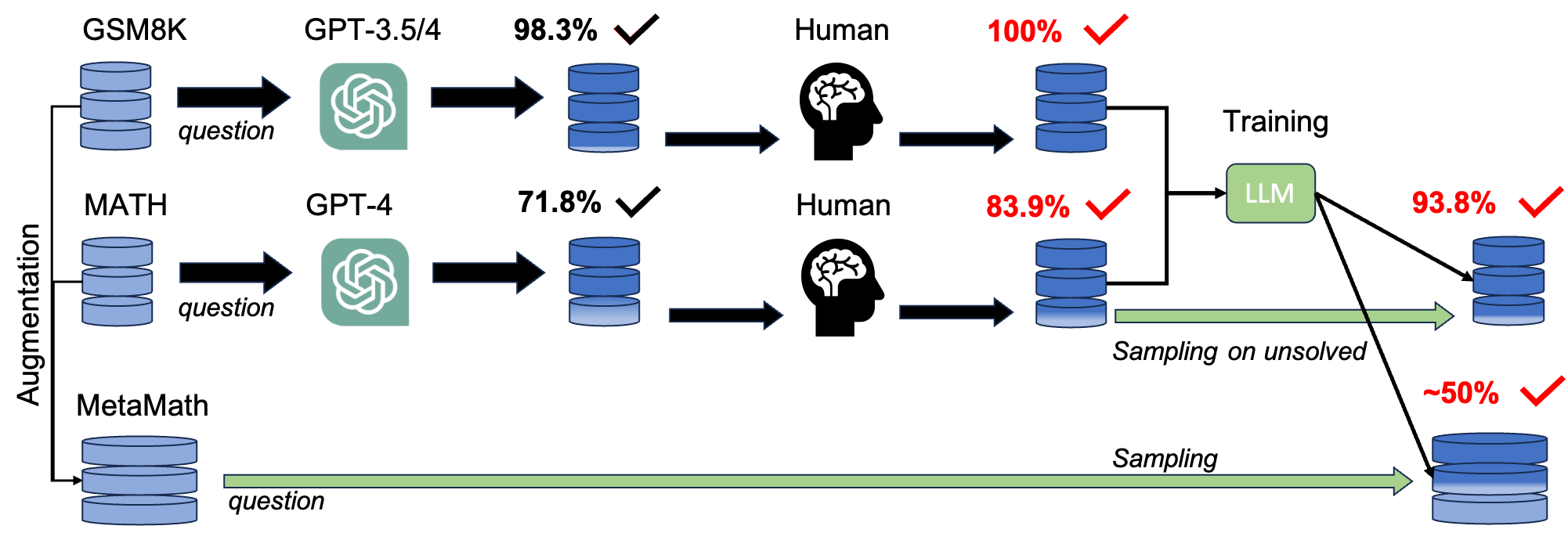}
    \caption{
    The data pipeline illustrates the process of data generation for the GSM8K and MATH datasets. 
    We employ GPT-4 to provide initial annotations, followed by human verification to fix easily rectifiable errors. 
    For MATH dataset, an additional sampling strategy derived from a self-trained Large Language Model (LLM) is imposed. 
    }
    \label{fig:data}
\end{figure}

\textbf{MATH} The foundational approach to constructing solutions for the MATH dataset continues to rely on the REACT instruction, with the specifics of the prompt setup available for reference in Appendix~\ref{app:data}. 
However, two primary distinctions exist. 
First, in the initial four attempts employing GPT, we exclusively utilize GPT-4 with maximum 8 code snippets allowed due to the substantially higher difficulty of the MATH questions. 
Following this process, a mere 71.8\% of the questions in the MATH training set are provided with at least one correct solution.

Secondly, we forgo manual solutions for all remaining 1,208 questions due to the significant burden it imposes, requiring both specialized knowledge and an excessive budget. 
Instead, human verification is applied for the correction of easily rectifiable errors. 
Consequently, 83.9\% of MATH questions are correctly solved, then combined with previously created GSM8K dataset into a unified dataset with size 26.9K. 
This was fine-tuned on a well-pretrained MATH LLM, Llemma-34B~\cite{azerbayev2023llemma}. 
Using the obtained teacher model, we generated up to 100 samples with a temperature of 0.6 for each unsolved question, continuing until at most 4 correct solution were identified. 
Through this approach, we achieved a coverage of 93.8\% of the questions in the MATH training set with correct solutions.

\textbf{Augmentation} To further augment the scale and variety of questions in our dataset, we incorporate about 240K novel questions sourced from MetaMath~\cite{yu2023metamath}. 
These questions, which are transformations of those found in the original GSM8K and MATH datasets, allow us to gather a richer set of sampled solutions including both positive and negative examples. 
We anticipate that this strategy of data augmentation will significantly boost the model's performance and serve the training signals for outcome value model.

\textbf{Re-formatting} The data generation pipeline for REACT is depicted in Figure~\ref{fig:data}, resulting in a consolidated dataset comprising approximately 28.8K solutions. 
However, the data we obtained using the REACT template's keywords was not used directly. 
Instead, we converted the REACT data into an HTML-like format, using \texttt{<p></p>} to encapsulate text analyses and \texttt{<code></code>} to encapsulate code snippets, as demonstrated in the Figure~\ref{fig:reformat} in Appendix~\ref{app:reformat}. 
We observed that employing REACT instructions typically yielded higher quality output from GPT models. 
Yet, when it comes to fine-tuning a pre-trained Large Language Model (LLM), utilizing an HTML-like format for the initial iteration results in a training loss that is, on average, approximately 20\% lower.
A thorough comparison is available in Section~\ref{sec:exp}. 
Our hypothesis is that the HTML-like format may bridge the gap between the pre-training corpus and the fine-tuning corpus, leading to improved initialization performance.

\section{Fine-Tuning}
\label{sec:ft}

To enhance a large language model's mathematical reasoning capabilities, we propose utilizing the foundational model Llemma~\cite{azerbayev2023llemma}. 
There are two primary reasons for choosing Llemma. 
First, Llemma represents a continuation of the pre-training process initiated by Llama-2~\cite{touvron2023llama}, extending its proficiency into both mathematical and coding domains, which aligns seamlessly with our requirements. 
Second, it has been demonstrated that neither Llama-2 nor Llemma exhibit excessive overfitting on the GSM8K or MATH dataset, as confirmed by \cite{wei2023skywork}.

\begin{figure}[t]
    \centering
    \includegraphics[width=\textwidth]{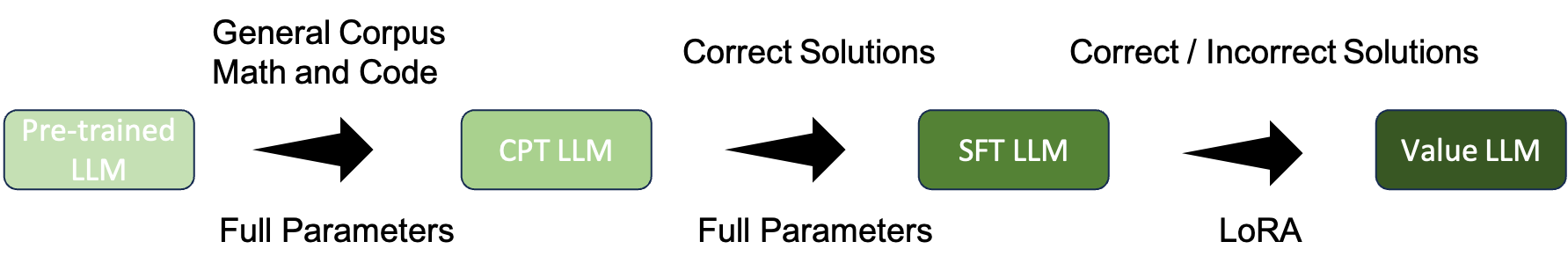}
    \caption{
    The training pipeline is divided into three distinct stages. In the initial stage, we continue pre-training using a comprehensive corpus that encompasses both mathematical and coding domains, e.g., Llemma~\cite{azerbayev2023llemma}. The second stage involves supervised fine-tuning in full parameters, utilizing our specially curated dataset. Finally, the third stage consists of multi-task supervised fine-tuning in LoRA mode, where the model is trained on a mixture of both correct and incorrect solutions.
    }
    \label{fig:ft}
\end{figure}

\subsection{Supervised Fine-Tuning with Full Parameters}

The supervised fine-tuning closely mirrors the data generation step that serves to expand the coverage of the MATH dataset. 
During the SFT stage, we tune the entire set of parameters of the LLM using our specially curated dataset. 
For each given question $\mathbf{q}$ and its corresponding correct solution $\mathbf{s}^+$, we optimize the model by minimizing the cross-entropy loss by masking out the tokens of questions.
\begin{equation}
   \min - \log p(\mathbf{s}^+|\mathbf{q})
\end{equation}

\subsection{Multi-Task Fine-Tuning with LoRA}

During solution generation, the Language Model (LLM) is capable of effortlessly creating both correct (positive) and incorrect (negative) samples $\mathbf{s}^-$. 
This duality enables us to train the LLM to discern the validity of a solution by predicting whether the final answer is accurate. 
To achieve this, we add a light-weighted binary classifier (\emph{i.e.}, a linear layer parameterized by $W\in \mathbb{R}^{d\times 1}$ and $b\in\mathbb{R}$ with sigmoid activation), alongside the existing softmax layer responsible for token prediction. 
We maintain a roughly equal ratio of positive to negative examples for balanced training. 
The overall loss follows the multi-task setting in \cite{cobbe2021training}, while the main difference is that our LLM is fine-tuned using the efficient training method known as LoRA~\cite{hu2021lora}.
\begin{equation}
   \min - \log p\left( \mathbf{s}^{+} | \mathbf{q} \right) - \log p\left( \mathbf{y}^{+} | (\mathbf{q}, \mathbf{s}^{+}) \right) - \log p\left( \mathbf{y}^{-} | (\mathbf{q}, \mathbf{s}^{-}) \right) 
\end{equation}
where the first term represents the traditional generation loss of training LLM on correct solutions, while the latter two terms correspond to the classification loss calculated for each token of either correct or incorrect solution. 
Therefore, $\mathbf{y}$ is a binary vector whose length is equal to the number of tokens in the solution $\mathbf{s}$. 
Similar to \cite{cobbe2021training}, the label of each token is completely determined by the correctness of solution, \emph{i.e.}, $\mathbf{y}^+=\mathbf{1}$; and $\mathbf{y}^-=\mathbf{0}$.

Given the significantly larger data size required for training the Value LLM, we employ LoRA during fine-tuning to be more computationally efficient. 
In addition, the LoRA training also retrains the model's generative abilities, \emph{i.e.}, playing a dual role on generation solutions and evaluating them. 
The primary benefits of this feature is practicality, as it requires the deployment of only a single LLM for the entire inference process. 
This ability, maintained along with the prediction of token-level values, allows for straightforward modifications to the decoding algorithm used in the transformer decoder implementation. 
For instance, the beam search mechanism could combine the original log-likelihood with the predicted value. 
This potential direction is what we intend to explore in future work.

\subsection{Inference} 

Both SFT LLM and Value LLM are capable of directly generating solutions. 
The solution-generating LLM can be paired with the Value LLM, which serves as an outcome value model (OVM). 
The Value LLM primarily serves to evaluate outcomes, that is, to estimate the likelihood of the final answer being correct. 
To improve the quality of solutions generated by the SFT LLM, one might consider re-ranking multiple solutions sampled from the SFT LLM. 
However, we suggest employing an outlier-free OVM selection algorithm to identify the best answer. 
Specifically, given $K$ sampled solutions $\{\mathbf{s_i}\}_{i=1}^K$ resulting in $k$ distinct final answers $\{\mathbf{a}_j\}_{j=1}^k$, the frequency of each answer is represented as $n_1, \cdots, n_k$, such that $\sum_{j=1}^k n_j = K$. 
The optimal answer is selected according to the criterion $\kappa = \arg\max_{\{j|n_j > \delta_K \}} \max_{\mathbf{s}_i \in \mathbf{a}_j} \text{OVM}(\mathbf{s}_{i})$. 
In our experiments, for $K = 20$, we set $\delta_K = 1$. 
This is because a small number of samples, such as 20 in our example, might lead to a situation where a random sample yields an anomalously high outcome prediction, making it crucial to exclude outlier solutions. 
In the rare case that all $K$ sampled solutions are unique, we simply choose the solution with the highest predicted outcome value.

\section{Experiments}
\label{sec:exp}

\subsection{Dataset Recap}
\label{sec:data_recap}

We present the statistics for our positive examples used in supervised fine-tuning in Table~\ref{tab:data_stat}. 
The seed data, derived from GSM8K and MATH datasets, culminates in a collection of 26.9K solutions by GPT and human annotations, which is mainly used to train our teacher model from Llemma-34B. 
For the augmentation data obtained from MetaMath, which encompasses 240K new questions, we employed the teacher model to sample one solution for each question and randomly select approximately 55K question-correct solution pairs. 
When applying self-generation to create positive examples, a significant number of incorrect solutions are inevitably produced. 
These can serve as training data for the outcome value model. 
In total, we have gathered 300K examples, both positive and negative, maintaining an approximately balanced ratio of labels.

The in-domain test sets come from the original GSM8K and MATH datasets. 
We also conduct evaluations on two out-of-domain (OOD) test sets: the open-source OCWCourses dataset~\cite{lewkowycz2022solving} and our proprietary GaoKao2023-Math-En dataset. 
OCWCourses comprises a collection of 272 STEM problems aimed at the undergraduate level, requiring multi-step reasoning for most questions. 
The GaoKao2023-Math-En dataset consists of 385 mathematics problems from the 2023 Chinese higher education entrance examination, professionally translated into English. 
The two OOD datasets are even greater challenging than MATH dataset.

\begin{table}[t]
\centering
\caption{Data statistics}
\label{tab:data_stat}
\begin{tabular}{l|c|c|c|c|c|c}
\toprule
Data source &  \multicolumn{3}{c|}{Generation method}  & Total   & \# Code & Trainset  \\
\cline{2-4}
            & GPT & Human & Self-train & solutions   & snippets  &  coverage \\ 
\hline
\multicolumn{7}{c}{SFT (correct solutions)}\\
\hline
GSM8K   & 17,480  & 95  &   -       & 17,576    & $\leq 5$ & 7,473 / 7,473 \\
MATH    &  6,483  & 2,862 & 1,933     & 11,277    & $\leq 8$ & 7,011 / 7,500 \\
\hline
MetaMath & -      & - & $\sim$55K & $\sim$55K &   $\leq 8$   & - \\
\hline
\multicolumn{7}{c}{OVM (correct / incorrect solutions)}\\
\hline
MetaMath & -      & - & $\sim$300K  & $\sim$300K  &   $\leq 8$   & - \\
\bottomrule
\end{tabular}
\end{table}

\begin{table}
    \centering
    \caption{Results on different datasets. The best results of open-source models are bold. $^*$GK2023-ME represents Gaokao-2023-Math-En dataset. $^\dagger$maj@K means majority voting over K samples. $^\S$The MARIO-OVM-7B here is simply used as an SFT LLM to generate one single solution. $^\ddagger$The MARIO-34B is the teacher model fine-tuned on 26.9K dataset for data generation.}
    \begin{tabular}{lccc|cc|cc}
      \toprule
      Model & Size & Tool & Zero & \multicolumn{2}{c|}{In-domain} & \multicolumn{2}{c}{Out-of-domain} \\
      \cline{5-8}
       &  &  & Shot & GSM8K & MATH & OCW & GK2023-ME$^*$ \\
      \midrule
      \multicolumn{8}{c}{Proprietary Models} \\
      \midrule
      GPT-4 & - & \XSolidBrush & \XSolidBrush & 92.0 & 42.5 & - & - \\
      GPT-4-Code & - & \checkmark & \XSolidBrush & 92.9 & 69.7 & - & - \\
      ChatGPT & - & \XSolidBrush & \XSolidBrush & 80.8 & 35.5 & - & - \\
      ChatGPT(PAL) & - & \checkmark & \XSolidBrush & 78.6 & 38.7 & - & - \\
      Claude-2 & - & \XSolidBrush & \XSolidBrush & 85.2 & 32.5 & - & - \\
      PaLM-2 & 540B & \XSolidBrush & \XSolidBrush & 80.7 & 34.3 & - & -\\
      Minerva & 540B & \XSolidBrush & \XSolidBrush & 58.8 & 33.6 & 17.6& -\\
      Gemini Ultra maj@32$^\dagger$ & - & \XSolidBrush & \XSolidBrush & 94.4 & 53.2 & - & -\\
      \midrule
      \multicolumn{8}{c}{Open-Source Models} \\
      \midrule
      Llama-2 & 7B & \XSolidBrush & \XSolidBrush & 13.3 & 4.1 & 3.7& -\\
      CodeLlama & 7B & \XSolidBrush & \XSolidBrush & 10.5 & 4.5 & 4.4 & -\\
      CodeLlama(PAL) & 7B & \checkmark & \XSolidBrush & 27.1 & 17.2 & - & -\\
      Llama-2 SFT & 7B & \XSolidBrush & \checkmark & 41.3 & 7.2 & - & -\\
      Llama-2 RFT & 7B & \XSolidBrush & \checkmark & 51.2 & - & - & -\\
      Llemma & 7B & \XSolidBrush & \XSolidBrush & 36.4 & 18.0 & 7.7& -\\
      Llemma(PAL) & 7B & \checkmark & \XSolidBrush & 40.1 & 21.5 & - & -\\
      Qwen & 7B & \XSolidBrush & \XSolidBrush & 51.7&	11.6 & - & - \\
      WizardMath & 7B & \XSolidBrush & \checkmark & 54.9 & 10.7 & - & -\\
      DeepSeek-Coder & 6.7B & \checkmark & \XSolidBrush & 43.2 & 19.2 & - & -\\
      MathCoder & 7B & \checkmark & \checkmark & 67.8 & 30.2 & - & -\\
      MAmmoTH-Coder & 7B & \checkmark & \XSolidBrush & 59.4 & 33.4 & 11.0 & 15.3\\
      ToRA & 7B & \checkmark & \checkmark & 68.8 & 40.1 & 2.6 & 19.5 \\
      ToRA-Code & 7B & \checkmark & \checkmark & 72.6 & 44.6 & 4.8 & 23.9\\
      \hline
      \textbf{MARIO} & 7B & \checkmark & \checkmark & 70.1 & 47.0 & \textbf{21.7} & \textbf{38.2} \\
      \textbf{MARIO-OVM-7B}$^\S$ & 7B & \checkmark & \checkmark & \textbf{74.5} & \textbf{48.3} & 21.0 & 34.8 \\
      \midrule
      CodeLlama & 34B & \XSolidBrush & \XSolidBrush & 29.6 & 12.2 & 7.0 & -\\
      CodeLlama(PAL) & 34B & \checkmark & \XSolidBrush & 53.3 & 23.9 & - & -\\
      Llemma & 34B & \XSolidBrush & \XSolidBrush & 51.5 & 25.0 & 11.8& -\\
      Llemma(PAL) & 34B & \checkmark & \XSolidBrush & 62.6 & 27.1 & - & -\\
      DeepSeek-Coder & 33B & \checkmark & \XSolidBrush & 60.7 & 29.1 & - & -\\
      MathCoder & 34B & \checkmark & \checkmark & \textbf{81.7} & 45.2 & - & -\\
      MAmmoTH-Coder & 34B & \checkmark & \XSolidBrush & 72.7 & 43.6 & 14.0 & 25.2\\
      ToRA-Code & 34B & \checkmark & \checkmark & 80.7 & 50.8 & 5.5 & 31.7 \\
      \hline
      \textbf{MARIO}$^\ddagger$ & 34B & \checkmark & \checkmark & 78.2 & \textbf{53.5} & \textbf{30.2} & \textbf{42.6} \\
      \midrule
      DeepSeek-Chat & 67B & \XSolidBrush & \XSolidBrush & 84.1 & 32.6 & - & - \\
      WizardMath & 70B & \XSolidBrush & \checkmark & 81.6& 22.7& - & -\\
      MathCoder & 70B & \checkmark & \checkmark & 83.9 & 45.1 & - & -\\
      MAmmoTH & 70B & \checkmark & \XSolidBrush & 76.9 & 41.8 & 11.8 & 24.7\\
      ToRA & 70B & \checkmark & \checkmark & 84.3& 49.7& 9.6 & 30.9\\
      Qwen & 72B & \XSolidBrush & \XSolidBrush & 78.9 & 35.2 & - & - \\
      \bottomrule
    \end{tabular}
    \label{tab:main_exp}
\end{table}

\subsection{Implementation Details}

We train the Llemma series~\cite{azerbayev2023llemma} through fine-tuning with our curated corpus, resulting in the development of our SFT LLM series, because, on GSM8K and MATH, Llama-2 was verified~\cite{wei2023skywork} to have no data leakage issue, and its CPT model Llemma had undergone data overlapping checking~\cite{azerbayev2023llemma}. 
During this optimization phase, we generally employed a learning rate of 5e-5, with the exception of the 7B and 34B models, for which we reduced the rate to 4e-5. 
We set the global batch size at 512 and used a linear learning rate scheduler that included a warm-up phase constituting 3\% of the total training duration, spread over 3 epochs. 
Training for all models was launched with the accelerate\footnote{\url{https://github.com/huggingface/accelerate}} in DeepSpeed ZeRO Stage2~\cite{deepspeedzero} and Flash-Attention 2 mechanism~\cite{dao2023flashattention2}.

When fine-tuning the value LLM with LoRA, we configure the hyper-parameters with a rank of 4096 and an alpha of 2048 for the attention parameters $W_q$ and $W_v$. 
In the context of the Llama-2-7B architecture, 2B model parameters are trainable. 
We employ a learning rate of 5e-5, which is progressively adjusted using a cosine decay scheduler.

During inference, the maximum number of code snippets permitted remains consistent with the prepared training data, with a limit of 5 for GSM8K and 8 for MATH. 
For the more challenging out-of-domain datasets, the number is also set at 8 code snippets. 

\textbf{Baselines} We conducted comparisons with renowned proprietary and open-source pretrained LLMs such as GPT~\cite{openai2023gpt}, Claude~\cite{claude}, PaLM~\cite{anil2023palm}, Minerva~\cite{lewkowycz2022solving}, Gemini~\cite{team2023gemini}, Llama-2~\cite{touvron2023llama}, CodeLlama~\cite{roziere2023code}, Qwen~\cite{bai2023qwen}, and DeepSeek~\cite{deepseek-coder}. 
Additionally, we have reported results from a variety of open-source models, most notably Llama-2, along with several SFT models derived from Llama-2, including RFT~\cite{yuan2023scaling}, WizardMath~\cite{luo2023wizardmath}, MathCoder~\cite{wang2023mathcoder}, MAmmoTH~\cite{yue2023mammoth} and ToRA~\cite{gou2023tora}.

\subsection{Main Results}
\label{sec:exp:main}

\textbf{SFT Model} Table \ref{tab:main_exp} demonstrates the performance of our model across four datasets that encompass both in-domain and out-of-domain problems. 
Our 34B teacher model, MARIO-34B, which is fine-tuned exclusively on the 26.9K dataset, does not match the performance of MathCoder or ToRA at the grade school math level. 
In contrast, for more complex problems in the MATH dataset, or even for challenging out-of-domain problems, our 34B model consistently outperforms others. 
Fewer training data may be one reason, but the main reason should be its capability to perform text analysis, which breaks down problems into manageable code snippets, thus enhancing its problem-solving effectiveness. 
This is verified by the similar pattern observed with our 7B model that was trained on an 82K dataset, which is in line with past state-of-the-art (SOTA) methods. 
So we conclude that such a model achieves superior results on more complex problems, likely because these problems demand more than simple logic and a few arithmetic steps — scenarios where models with a code-centric approach typically have an edge.

\textbf{OVM Model} The experimental findings of our outlier-free OVM selection algorithm are displayed in Table~\ref{tab:ovm}, where we contrast our approach with the majority voting algorithm. 
Our findings indicate that the outlier-free OVM algorithm can significantly enhance the results compared to majority voting, and the gain of majority voting by our approach is more significant than ToRA. 
As previously mentioned, the OVM model is designed for multi-task learning, enabling it to not only evaluate outcomes but also generate solutions. 
In Table~\ref{tab:ovm}, we present a comprehensive results showcasing the OVM's performance when it takes on both the roles of solution generation and outcome evaluation. 
The OVM demonstrates a comparable proficiency in generating solutions; however, it exhibits a slightly reduced effectiveness on out-of-domain datasets. 
This outcome is to be expected, given that our OVM has been continually fine-tuned on the MetaMath questions, originating from the GSM8K and MATH datasets.

\begin{table}[t]
\centering
    \caption{Results on OVM-7B.}
    \label{tab:ovm}
    \begin{tabular}{l|l|l|l|l}
    \toprule
    Inference method & GSM8K & MATH & OCWCourses & GaoKao2023-Math-En\\
    \hline
    ToRA-Code-7B        & 72.6 & 44.6 & 4.8  & 23.9 \\ 
    ~~+maj@\textcolor{blue}{50}           & 76.8 \textcolor{red}{+4.2} & 52.5 \textcolor{red}{+7.9} & -    & -    \\
    \hline
    MARIO-7B            & 70.1 & 47.0 & 21.7 & 38.2 \\
    ~~+maj@\textcolor{cyan}{20}           & 80.5 \textcolor{red}{+10.4} & 56.7 \textcolor{red}{+9.7} & 25.4 \textcolor{red}{+3.7} & 41.6 \textcolor{red}{+3.4} \\
    ~~~~+OVM-7B@\textcolor{cyan}{20}      & 82.9 \textcolor{red}{+12.8} & 59.1 \textcolor{red}{+12.1} & \textbf{28.3} \textcolor{red}{+6.6} & \textbf{45.2} \textcolor{red}{+7.0} \\
    \hline
    MARIO-OVM-7B        & 74.5 & 48.3 & 21.0 & 34.8 \\
    ~~+maj@\textcolor{cyan}{20}          & \textbf{83.8} \textcolor{red}{+9.3} & 59.7 \textcolor{red}{+11.4} & 22.1 \textcolor{red}{+1.1} & 43.6 \textcolor{red}{+8.8} \\
    ~~~~+OVM-7B@\textcolor{cyan}{20}      & 83.6 \textcolor{red}{+9.1} & \textbf{60.6} \textcolor{red}{+12.3} & 25.4 \textcolor{red}{+4.4} & 42.9 \textcolor{red}{+8.1} \\
    \bottomrule
    \end{tabular}
\end{table}

\subsection{Ablation Study}

We perform the first ablation study to examines the impact of each data source by incrementally adding more training examples, with the primary findings detailed in Table~\ref{tab:ablation}. 
Overall, the advancements in MATH are more pronounced. 
We ascribe this trend to three main factors. 
First, the GSM8K dataset, synthesized by GPT, encompasses 98.3\% of the questions, in contrast to the MATH dataset's 71.8\% coverage. 
Secondly, the selection criterion for the MATH dataset hinges on an exact match between GPT's generated final answer and the dataset's provided answer, which could lead the fine-tuned model to overfit specific questions that have straightforward answers. 
Human intervention has the potential to enhance the variability of the answers. 
Lastly, the teacher model's generated solutions concentrate more heavily on the MATH dataset. 
This is due to the fact that we have extracted a greater number of multi-step reasoning solutions according to MetaMath questions, which are likely better suited to the difficulty of the MATH dataset.

The second ablation study aims to investigate the impact of the foundational math LLM and the data formatting for SFT. 
DeepSeek-MATH-7B~\cite{shao2024deepseekmath}, is a specialized math-focused LLM developed through continual pre-training on the Deep-Seek-Code-7B model, which benefits from a more extensive math pre-training corpus than Llemma-7B and purposely omits any content that may relate to GSM8K and MATH datasets. 
Consequently, DeepSeek-MATH-7B is supposed to outperform Llemma-7B. 
When applying SFT on a large scale code-centric SFT dataset, DeepSeek-MATH-7B can achieve the SOTA performance as 7B LLM. 
The result presented in the second row of Table~\ref{tab:ablation2} shows the results of fine-tuning DeepSeek-MATH-7B with our dataset. 
Despite being only 1/30th the size of their used dataset, our hybrid format demonstrates greater data efficiency. 
The result presented in the last row of Table~\ref{tab:ablation2} suggest that SFT from a superior CPT model enhances math reasoning capabilities.

\begin{table}[t]
\centering
\caption{Ablation study of data usage on 7B model.}
\label{tab:ablation}
\begin{tabular}{l|c|ll}
\toprule
Data used & \# trainset & GSM8K & MATH \\
\hline
GPT                & 23.9K & 66.3 & 40.2 \\
~+Human            & 26.9K & 67.1 \textcolor{red}{+0.8} & 43.5 \textcolor{red}{+3.3} \\
~~+MATH Aug        & 28.8K & 67.4 \textcolor{red}{+1.1} & 44.4 \textcolor{red}{+4.2} \\
~~~+MetaMath Aug   & 82K   & 70.1 \textcolor{red}{+3.8} & 47.0 \textcolor{red}{+6.8} \\
\bottomrule
\end{tabular}
\end{table}

\begin{table}[t]
\centering
\caption{Ablation study of CPT Model and Data format. $^\#$The result is sourced from~\cite{shao2024deepseekmath}.}
\label{tab:ablation2}
\begin{tabular}{l|c|c|cc}
\toprule
CPT Model & data size & data format & GSM8K & MATH.\\
\hline
DeepSeek-Math-7B$^\#$   & 776K & code & 83.7 & 57.4 \\
\hline
DeepSeek-Math-7B   & 28.8K & text+code& 78.4 & 56.1 \\ 
Llemma-Math-7B     & 28.8K & text+code & 67.4 & 44.4 \\
\bottomrule
\end{tabular}
\end{table}

\subsection{Why GSM-Hard is not a good testset for MATH LLM?}

The GSM-Hard dataset, introduced by \citet{gao2023pal}, is akin to the original GSM8K test set, with the sole distinction being the alteration of numbers in the original problem statements. 
However, as illustrated in the introduction, these modifications to the numbers do not always align with the common sense of real physical world; for example, ages cannot be negative, and the number of people cannot be fractional. 
Methods following the PAL or PoT paradigm tend to generate code without verifying the sensibility of their output. 
In contrast, our approach incorporates a textual analysis that ensures the results derived from code execution are consistent with the constraints of the physical world. 
As a result, our LLM will opt not to produce an illogical final answer or to arbitrarily round fractions, even if the so-called correct answer has been computed during the problem-solving process. 
This accounts for the lower accuracy observed with our method on this dataset, as shown in Table~\ref{tab:gsm_hard}. 
In addition, we found the solutions in the GSM-Hard remain the same as the original GSM8K, even the numbers have changed. 
For more intricate examples, please refer to the case study provided in Appendix~\ref{app:gsm_hard}. 
In summary, we suggest not the use GSM-Hard dataset in this community unless the mentioned errors have been fixed.

\begin{table}[h]
    \centering
    \caption{Results on GSM-Hard~\cite{gao2023pal}. $^*$PAL is based on code-davinci-002.}
    \label{tab:gsm_hard}
    \begin{tabular}{l|cccccc}
    \toprule
    Model       & PAL~\cite{gao2023pal}  & DeepSeek-Coder & MAmmoTH & ToRA-Code & MARIO & MARIO-OVM\\
    \hline
    Size        & 175B$^*$ & 6.7B        & 7B    & 7B    & 7B   & 7B\\
    \hline    
    Accuracy    & 61.2     & 40.3        & 56.5  & 56.0  & 50.0 & 53.2  \\ 
    \bottomrule
    \end{tabular}
\end{table}


\subsection{Why re-formatting to HTML?}

The standard template for REACT examplified in Appendix~\ref{app:prompts} employs a key-value pair format represented as strings, with each step comprising elements like ``\texttt{Thought:text analysis}", ``\texttt{Action: tool name}", ``\texttt{Action Input: code snippet}", and ``\texttt{Observation: execution output}". 

\begin{wrapfigure}{r}{0.5\textwidth}
    \centering
    \includegraphics[width=0.48\textwidth,scale=1.00]{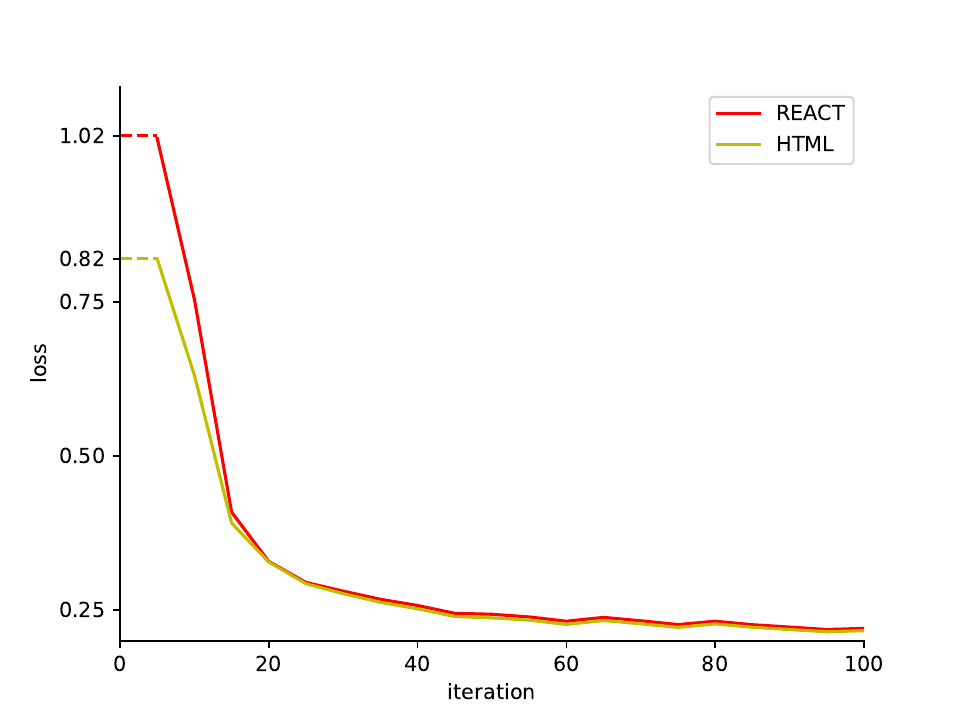}
    \caption{Train loss of different formats within first 100 iterations when fine-tuning on Llemma-34B.}
    \label{fig:format}
\end{wrapfigure}

REACT~\cite{yao2022react} is an excellent approach to empower the LLM the capability of using external tool. 
However, our analysis of the log-likelihood for REACT examples suggests that such data formats are rare in the corpora used for pre-training LLMs. 
In contrast, when converting REACT data into an HTML format, using tags such as ``\texttt{<p>text analysis</p>}" and ``\texttt{<code>code snippet</code>}", we noted a substantial decrease in log-likelihood. 
In our data pipeline of Figure~\ref{fig:data}, when training the Llemma-34B on the 26.9K dataset with different formats, this reduction was evident from the initial loss observed in the first iteration. 

As Figure~\ref{fig:format} demonstrates, the loss for fine-tuning with HTML-formatted data is 20\% lower than when using the original REACT format. 
However, upon evaluating the performance on the test sets after 3 epochs of SFT, we observed only a marginal improvement (~1\%) on MATH test set.   
Given the substantially reduced initial loss, we hypothesize that careful hyper-parameter tuning could potentially enhance the performance of data formatted in HTML. 
We intend to explore this possibility in future work.


\section{Related Works}
\label{sec:related}

\textbf{Mathematical reasoning} attracts more attentions because of the emergence of LLMs. 
Recent works~\cite{wei2022chain,kojima2023large,wang2023selfconsistency,deepseek-coder} on mathematical reasoning have made impressive progress empowered by LLMs. 
Yet exact calculations and symbolic manipulations within the reasoning process remain challenging. 
Some works have explored tools including calculators~\cite{cobbe2021training,shao-etal-2022-chaining} and code interpreters~\cite{gao2023pal} to address the limitations. 
Further research~\cite{wang2023mathcoder,yue2023mammoth,gou2023tora} attempt to combine tool-use and textual reasoning process to leverage the strengths of both.

\textbf{Knowledge distillation}~\cite{hinton2015distilling,Gou_2021} is a commonly used approach to promote student models by transferring knowledge from teacher models to them. 
Utilizing teacher LLM to construct reasoning samples for student model to fine-tune proved to be effective practice of knowledge distillation\cite{pmlr-v202-fu23d,ho-etal-2023-large}. 
Our corpus construction includes knowledge distillation of this kind on MATH with more samplings from 34B SFT LLM.

\textbf{Verification in mathematical reasoning} plays a crucial role in ensuring inference performance by allowing auto-regressive models to correct already-made errors. 
It has been proved that LLMs can self-verify~\cite{anonymous2023solving,weng-etal-2023-large,xie2023decomposition} and self-refine~\cite{madaan2023selfrefine} by designed prompting. 
A specifically trained verifier can also play a similar role by intervening the decoding process~\cite{cobbe2021training,khalifa-etal-2023-grace,yu2023outcome}. 
In this paper, we use multi-task fine-tuning which is similar to the training of a simple outcome supervision model.

\section{Conclusion}
\label{sec:concl}

This paper introduces a reproducible pipeline that covers both the construction of a math-specific dataset and the fine-tuning of a large language model (LLM) with an emphasis on leveraging a Python code interpreter. 
To support further research and development in this field, we first have made our model checkpoints available as an open-source resource and provide a detailed description of our three-stage fine-tuning process. 
Our results demonstrate that integrating text analyses with code snippets enhances the model's capabilities for common sense reasoning and precise computation in mathematical reasoning tasks. 
Moreover, our fine-tuning method enhances model performance by incorporating a verifier model that requires only a negligible number of additional parameters. 
To the best of our knowledge, our approach sets a new state-of-the-art benchmark for LLMs with a size around 7 billion parameters on the MATH datasets, and it exhibits notable generalization ability on challenging out-of-domain math datasets.

\newpage

\bibliography{neurips}

\begin{thebibliography}{44}
\providecommand{\natexlab}[1]{#1}
\providecommand{\url}[1]{\texttt{#1}}
\expandafter\ifx\csname urlstyle\endcsname\relax
  \providecommand{\doi}[1]{doi: #1}\else
  \providecommand{\doi}{doi: \begingroup \urlstyle{rm}\Url}\fi

\bibitem[Anil et~al.(2023)Anil, Dai, Firat, Johnson, Lepikhin, Passos, Shakeri,
  Taropa, Bailey, Chen, et~al.]{anil2023palm}
Rohan Anil, Andrew~M Dai, Orhan Firat, Melvin Johnson, Dmitry Lepikhin,
  Alexandre Passos, Siamak Shakeri, Emanuel Taropa, Paige Bailey, Zhifeng Chen,
  et~al.
\newblock Palm 2 technical report.
\newblock \emph{arXiv preprint arXiv:2305.10403}, 2023.

\bibitem[Anonymous(2023)]{anonymous2023solving}
Anonymous.
\newblock Solving challenging math word problems using {GPT}-4 code interpreter
  with code-based self-verification.
\newblock In \emph{Submitted to The Twelfth International Conference on
  Learning Representations}, 2023.
\newblock URL \url{https://openreview.net/forum?id=c8McWs4Av0}.
\newblock under review.

\bibitem[Anthropic(2023)]{claude}
Anthropic.
\newblock Model card and evaluations for claude models.
\newblock July 2023.

\bibitem[Azerbayev et~al.(2023)Azerbayev, Schoelkopf, Paster, Santos, McAleer,
  Jiang, Deng, Biderman, and Welleck]{azerbayev2023llemma}
Zhangir Azerbayev, Hailey Schoelkopf, Keiran Paster, Marco~Dos Santos, Stephen
  McAleer, Albert~Q Jiang, Jia Deng, Stella Biderman, and Sean Welleck.
\newblock Llemma: An open language model for mathematics.
\newblock \emph{arXiv preprint arXiv:2310.10631}, 2023.

\bibitem[Bai et~al.(2023)Bai, Bai, Chu, Cui, Dang, Deng, Fan, Ge, Han, Huang,
  Hui, Ji, Li, Lin, Lin, Liu, Liu, Lu, Lu, Ma, Men, Ren, Ren, Tan, Tan, Tu,
  Wang, Wang, Wang, Wu, Xu, Xu, Yang, Yang, Yang, Yang, Yao, Yu, Yuan, Yuan,
  Zhang, Zhang, Zhang, Zhang, Zhou, Zhou, Zhou, and Zhu]{bai2023qwen}
Jinze Bai, Shuai Bai, Yunfei Chu, Zeyu Cui, Kai Dang, Xiaodong Deng, Yang Fan,
  Wenbin Ge, Yu~Han, Fei Huang, Binyuan Hui, Luo Ji, Mei Li, Junyang Lin, Runji
  Lin, Dayiheng Liu, Gao Liu, Chengqiang Lu, Keming Lu, Jianxin Ma, Rui Men,
  Xingzhang Ren, Xuancheng Ren, Chuanqi Tan, Sinan Tan, Jianhong Tu, Peng Wang,
  Shijie Wang, Wei Wang, Shengguang Wu, Benfeng Xu, Jin Xu, An~Yang, Hao Yang,
  Jian Yang, Shusheng Yang, Yang Yao, Bowen Yu, Hongyi Yuan, Zheng Yuan,
  Jianwei Zhang, Xingxuan Zhang, Yichang Zhang, Zhenru Zhang, Chang Zhou,
  Jingren Zhou, Xiaohuan Zhou, and Tianhang Zhu.
\newblock Qwen technical report, 2023.

\bibitem[Chen et~al.(2022)Chen, Ma, Wang, and Cohen]{chen2022program}
Wenhu Chen, Xueguang Ma, Xinyi Wang, and William~W Cohen.
\newblock Program of thoughts prompting: Disentangling computation from
  reasoning for numerical reasoning tasks.
\newblock \emph{arXiv preprint arXiv:2211.12588}, 2022.

\bibitem[Cobbe et~al.(2021{\natexlab{a}})Cobbe, Kosaraju, Bavarian, Chen, Jun,
  Kaiser, Plappert, Tworek, Hilton, Nakano, Hesse, and
  Schulman]{gsm8kcobbe2021}
Karl Cobbe, Vineet Kosaraju, Mohammad Bavarian, Mark Chen, Heewoo Jun, Lukasz
  Kaiser, Matthias Plappert, Jerry Tworek, Jacob Hilton, Reiichiro Nakano,
  Christopher Hesse, and John Schulman.
\newblock Training verifiers to solve math word problems.
\newblock \emph{arXiv preprint arXiv:2110.14168}, 2021{\natexlab{a}}.

\bibitem[Cobbe et~al.(2021{\natexlab{b}})Cobbe, Kosaraju, Bavarian, Chen, Jun,
  Kaiser, Plappert, Tworek, Hilton, Nakano, et~al.]{cobbe2021training}
Karl Cobbe, Vineet Kosaraju, Mohammad Bavarian, Mark Chen, Heewoo Jun, Lukasz
  Kaiser, Matthias Plappert, Jerry Tworek, Jacob Hilton, Reiichiro Nakano,
  et~al.
\newblock Training verifiers to solve math word problems.
\newblock \emph{arXiv preprint arXiv:2110.14168}, 2021{\natexlab{b}}.

\bibitem[Dao(2023)]{dao2023flashattention2}
Tri Dao.
\newblock Flash{A}ttention-2: Faster attention with better parallelism and work
  partitioning.
\newblock 2023.

\bibitem[DeepSeek(2023)]{deepseek-coder}
DeepSeek.
\newblock Deepseek coder: Let the code write itself.
\newblock \url{https://github.com/deepseek-ai/DeepSeek-Coder}, 2023.

\bibitem[Fu et~al.(2023)Fu, Peng, Ou, Sabharwal, and Khot]{pmlr-v202-fu23d}
Yao Fu, Hao Peng, Litu Ou, Ashish Sabharwal, and Tushar Khot.
\newblock Specializing smaller language models towards multi-step reasoning.
\newblock In Andreas Krause, Emma Brunskill, Kyunghyun Cho, Barbara Engelhardt,
  Sivan Sabato, and Jonathan Scarlett, editors, \emph{Proceedings of the 40th
  International Conference on Machine Learning}, volume 202 of
  \emph{Proceedings of Machine Learning Research}, pages 10421--10430. PMLR,
  23--29 Jul 2023.
\newblock URL \url{https://proceedings.mlr.press/v202/fu23d.html}.

\bibitem[Gao et~al.(2023)Gao, Madaan, Zhou, Alon, Liu, Yang, Callan, and
  Neubig]{gao2023pal}
Luyu Gao, Aman Madaan, Shuyan Zhou, Uri Alon, Pengfei Liu, Yiming Yang, Jamie
  Callan, and Graham Neubig.
\newblock Pal: Program-aided language models.
\newblock In \emph{International Conference on Machine Learning}, pages
  10764--10799. PMLR, 2023.

\bibitem[Gou et~al.(2021)Gou, Yu, Maybank, and Tao]{Gou_2021}
Jianping Gou, Baosheng Yu, Stephen~J. Maybank, and Dacheng Tao.
\newblock Knowledge distillation: A survey.
\newblock \emph{International Journal of Computer Vision}, 129\penalty0
  (6):\penalty0 1789–1819, March 2021.
\newblock ISSN 1573-1405.
\newblock \doi{10.1007/s11263-021-01453-z}.
\newblock URL \url{http://dx.doi.org/10.1007/s11263-021-01453-z}.

\bibitem[Gou et~al.(2023)Gou, Shao, Gong, yelong shen, Yang, Huang, Duan, and
  Chen]{gou2023tora}
Zhibin Gou, Zhihong Shao, Yeyun Gong, yelong shen, Yujiu Yang, Minlie Huang,
  Nan Duan, and Weizhu Chen.
\newblock Tora: A tool-integrated reasoning agent for mathematical problem
  solving, 2023.

\bibitem[Hendrycks et~al.(2021)Hendrycks, Burns, Kadavath, Arora, Basart, Tang,
  Song, and Steinhardt]{mathhendrycks2021}
Dan Hendrycks, Collin Burns, Saurav Kadavath, Akul Arora, Steven Basart, Eric
  Tang, Dawn Song, and Jacob Steinhardt.
\newblock Measuring mathematical problem solving with the math dataset.
\newblock \emph{NeurIPS}, 2021.

\bibitem[Hinton et~al.(2015)Hinton, Vinyals, and Dean]{hinton2015distilling}
Geoffrey Hinton, Oriol Vinyals, and Jeff Dean.
\newblock Distilling the knowledge in a neural network, 2015.

\bibitem[Ho et~al.(2023)Ho, Schmid, and Yun]{ho-etal-2023-large}
Namgyu Ho, Laura Schmid, and Se-Young Yun.
\newblock Large language models are reasoning teachers.
\newblock In Anna Rogers, Jordan Boyd-Graber, and Naoaki Okazaki, editors,
  \emph{Proceedings of the 61st Annual Meeting of the Association for
  Computational Linguistics (Volume 1: Long Papers)}, pages 14852--14882,
  Toronto, Canada, July 2023. Association for Computational Linguistics.
\newblock \doi{10.18653/v1/2023.acl-long.830}.
\newblock URL \url{https://aclanthology.org/2023.acl-long.830}.

\bibitem[Hu et~al.(2021)Hu, Wallis, Allen-Zhu, Li, Wang, Wang, Chen,
  et~al.]{hu2021lora}
Edward~J Hu, Phillip Wallis, Zeyuan Allen-Zhu, Yuanzhi Li, Shean Wang, Lu~Wang,
  Weizhu Chen, et~al.
\newblock Lora: Low-rank adaptation of large language models.
\newblock In \emph{International Conference on Learning Representations}, 2021.

\bibitem[Khalifa et~al.(2023)Khalifa, Logeswaran, Lee, Lee, and
  Wang]{khalifa-etal-2023-grace}
Muhammad Khalifa, Lajanugen Logeswaran, Moontae Lee, Honglak Lee, and Lu~Wang.
\newblock {GRACE}: Discriminator-guided chain-of-thought reasoning.
\newblock In Houda Bouamor, Juan Pino, and Kalika Bali, editors, \emph{Findings
  of the Association for Computational Linguistics: EMNLP 2023}, pages
  15299--15328, Singapore, December 2023. Association for Computational
  Linguistics.
\newblock \doi{10.18653/v1/2023.findings-emnlp.1022}.
\newblock URL \url{https://aclanthology.org/2023.findings-emnlp.1022}.

\bibitem[Kojima et~al.(2023)Kojima, Gu, Reid, Matsuo, and
  Iwasawa]{kojima2023large}
Takeshi Kojima, Shixiang~Shane Gu, Machel Reid, Yutaka Matsuo, and Yusuke
  Iwasawa.
\newblock Large language models are zero-shot reasoners, 2023.

\bibitem[Lewkowycz et~al.(2022)Lewkowycz, Andreassen, Dohan, Dyer, Michalewski,
  Ramasesh, Slone, Anil, Schlag, Gutman-Solo, et~al.]{lewkowycz2022solving}
Aitor Lewkowycz, Anders Andreassen, David Dohan, Ethan Dyer, Henryk
  Michalewski, Vinay Ramasesh, Ambrose Slone, Cem Anil, Imanol Schlag, Theo
  Gutman-Solo, et~al.
\newblock Solving quantitative reasoning problems with language models.
\newblock \emph{Advances in Neural Information Processing Systems},
  35:\penalty0 3843--3857, 2022.

\bibitem[Lightman et~al.(2023)Lightman, Kosaraju, Burda, Edwards, Baker, Lee,
  Leike, Schulman, Sutskever, and Cobbe]{lightman2023let}
Hunter Lightman, Vineet Kosaraju, Yura Burda, Harri Edwards, Bowen Baker, Teddy
  Lee, Jan Leike, John Schulman, Ilya Sutskever, and Karl Cobbe.
\newblock Let's verify step by step.
\newblock \emph{arXiv preprint arXiv:2305.20050}, 2023.

\bibitem[Luo et~al.(2023)Luo, Sun, Xu, Zhao, Lou, Tao, Geng, Lin, Chen, and
  Zhang]{luo2023wizardmath}
Haipeng Luo, Qingfeng Sun, Can Xu, Pu~Zhao, Jianguang Lou, Chongyang Tao, Xiubo
  Geng, Qingwei Lin, Shifeng Chen, and Dongmei Zhang.
\newblock Wizardmath: Empowering mathematical reasoning for large language
  models via reinforced evol-instruct.
\newblock \emph{arXiv preprint arXiv:2308.09583}, 2023.

\bibitem[Madaan et~al.(2023)Madaan, Tandon, Gupta, Hallinan, Gao, Wiegreffe,
  Alon, Dziri, Prabhumoye, Yang, Welleck, Majumder, Gupta, Yazdanbakhsh, and
  Clark]{madaan2023selfrefine}
Aman Madaan, Niket Tandon, Prakhar Gupta, Skyler Hallinan, Luyu Gao, Sarah
  Wiegreffe, Uri Alon, Nouha Dziri, Shrimai Prabhumoye, Yiming Yang, Sean
  Welleck, Bodhisattwa~Prasad Majumder, Shashank Gupta, Amir Yazdanbakhsh, and
  Peter Clark.
\newblock Self-refine: Iterative refinement with self-feedback, 2023.

\bibitem[OpenAI(2023)]{openai2023gpt}
R~OpenAI.
\newblock Gpt-4 technical report.
\newblock \emph{arXiv}, pages 2303--08774, 2023.

\bibitem[Rajbhandari et~al.(2021)Rajbhandari, Ruwase, Rasley, Smith, and
  He]{deepspeedzero}
Samyam Rajbhandari, Olatunji Ruwase, Jeff Rasley, Shaden Smith, and Yuxiong He.
\newblock Zero-infinity: Breaking the gpu memory wall for extreme scale deep
  learning.
\newblock In \emph{Proceedings of the International Conference for High
  Performance Computing, Networking, Storage and Analysis}, SC '21, New York,
  NY, USA, 2021. Association for Computing Machinery.
\newblock ISBN 9781450384421.
\newblock \doi{10.1145/3458817.3476205}.
\newblock URL \url{https://doi.org/10.1145/3458817.3476205}.

\bibitem[Roziere et~al.(2023)Roziere, Gehring, Gloeckle, Sootla, Gat, Tan, Adi,
  Liu, Remez, Rapin, et~al.]{roziere2023code}
Baptiste Roziere, Jonas Gehring, Fabian Gloeckle, Sten Sootla, Itai Gat,
  Xiaoqing~Ellen Tan, Yossi Adi, Jingyu Liu, Tal Remez, J{\'e}r{\'e}my Rapin,
  et~al.
\newblock Code llama: Open foundation models for code.
\newblock \emph{arXiv preprint arXiv:2308.12950}, 2023.

\bibitem[Shao et~al.(2022)Shao, Huang, and Huang]{shao-etal-2022-chaining}
Zhihong Shao, Fei Huang, and Minlie Huang.
\newblock Chaining simultaneous thoughts for numerical reasoning.
\newblock In Yoav Goldberg, Zornitsa Kozareva, and Yue Zhang, editors,
  \emph{Findings of the Association for Computational Linguistics: EMNLP 2022},
  pages 2533--2547, Abu Dhabi, United Arab Emirates, December 2022. Association
  for Computational Linguistics.
\newblock \doi{10.18653/v1/2022.findings-emnlp.187}.
\newblock URL \url{https://aclanthology.org/2022.findings-emnlp.187}.

\bibitem[Shao et~al.(2024)Shao, Wang, Zhu, Xu, Song, Zhang, Li, Wu, and
  Guo]{shao2024deepseekmath}
Zhihong Shao, Peiyi Wang, Qihao Zhu, Runxin Xu, Junxiao Song, Mingchuan Zhang,
  YK~Li, Y~Wu, and Daya Guo.
\newblock Deepseekmath: Pushing the limits of mathematical reasoning in open
  language models.
\newblock \emph{arXiv preprint arXiv:2402.03300}, 2024.

\bibitem[Team et~al.(2023)Team, Anil, Borgeaud, Wu, Alayrac, Yu, Soricut,
  Schalkwyk, Dai, Hauth, et~al.]{team2023gemini}
Gemini Team, Rohan Anil, Sebastian Borgeaud, Yonghui Wu, Jean-Baptiste Alayrac,
  Jiahui Yu, Radu Soricut, Johan Schalkwyk, Andrew~M Dai, Anja Hauth, et~al.
\newblock Gemini: a family of highly capable multimodal models.
\newblock \emph{arXiv preprint arXiv:2312.11805}, 2023.

\bibitem[Touvron et~al.(2023)Touvron, Martin, Stone, Albert, Almahairi, Babaei,
  Bashlykov, Batra, Bhargava, Bhosale, et~al.]{touvron2023llama}
Hugo Touvron, Louis Martin, Kevin Stone, Peter Albert, Amjad Almahairi, Yasmine
  Babaei, Nikolay Bashlykov, Soumya Batra, Prajjwal Bhargava, Shruti Bhosale,
  et~al.
\newblock Llama 2: Open foundation and fine-tuned chat models.
\newblock \emph{arXiv preprint arXiv:2307.09288}, 2023.

\bibitem[Wang et~al.(2023{\natexlab{a}})Wang, Ren, Zhou, Lu, Luo, Shi, Zhang,
  Song, Zhan, and Li]{wang2023mathcoder}
Ke~Wang, Houxing Ren, Aojun Zhou, Zimu Lu, Sichun Luo, Weikang Shi, Renrui
  Zhang, Linqi Song, Mingjie Zhan, and Hongsheng Li.
\newblock Mathcoder: Seamless code integration in llms for enhanced
  mathematical reasoning, 2023{\natexlab{a}}.

\bibitem[Wang et~al.(2023{\natexlab{b}})Wang, Wei, Schuurmans, Le, Chi, Narang,
  Chowdhery, and Zhou]{wang2023selfconsistency}
Xuezhi Wang, Jason Wei, Dale Schuurmans, Quoc~V Le, Ed~H. Chi, Sharan Narang,
  Aakanksha Chowdhery, and Denny Zhou.
\newblock Self-consistency improves chain of thought reasoning in language
  models.
\newblock In \emph{The Eleventh International Conference on Learning
  Representations}, 2023{\natexlab{b}}.
\newblock URL \url{https://openreview.net/forum?id=1PL1NIMMrw}.

\bibitem[Wei et~al.(2022)Wei, Wang, Schuurmans, Bosma, Xia, Chi, Le, Zhou,
  et~al.]{wei2022chain}
Jason Wei, Xuezhi Wang, Dale Schuurmans, Maarten Bosma, Fei Xia, Ed~Chi, Quoc~V
  Le, Denny Zhou, et~al.
\newblock Chain-of-thought prompting elicits reasoning in large language
  models.
\newblock \emph{Advances in Neural Information Processing Systems},
  35:\penalty0 24824--24837, 2022.

\bibitem[Wei et~al.(2023)Wei, Zhao, Zhang, Zhu, Wang, Yang, Li, Cheng, L{\"u},
  Hu, et~al.]{wei2023skywork}
Tianwen Wei, Liang Zhao, Lichang Zhang, Bo~Zhu, Lijie Wang, Haihua Yang, Biye
  Li, Cheng Cheng, Weiwei L{\"u}, Rui Hu, et~al.
\newblock Skywork: A more open bilingual foundation model.
\newblock \emph{arXiv preprint arXiv:2310.19341}, 2023.

\bibitem[Weng et~al.(2023)Weng, Zhu, Xia, Li, He, Liu, Sun, Liu, and
  Zhao]{weng-etal-2023-large}
Yixuan Weng, Minjun Zhu, Fei Xia, Bin Li, Shizhu He, Shengping Liu, Bin Sun,
  Kang Liu, and Jun Zhao.
\newblock Large language models are better reasoners with self-verification.
\newblock In Houda Bouamor, Juan Pino, and Kalika Bali, editors, \emph{Findings
  of the Association for Computational Linguistics: EMNLP 2023}, pages
  2550--2575, Singapore, December 2023. Association for Computational
  Linguistics.
\newblock \doi{10.18653/v1/2023.findings-emnlp.167}.
\newblock URL \url{https://aclanthology.org/2023.findings-emnlp.167}.

\bibitem[Xie et~al.(2023)Xie, Kawaguchi, Zhao, Zhao, Kan, He, and
  Xie]{xie2023decomposition}
Yuxi Xie, Kenji Kawaguchi, Yiran Zhao, Xu~Zhao, Min-Yen Kan, Junxian He, and
  Qizhe Xie.
\newblock Decomposition enhances reasoning via self-evaluation guided decoding,
  2023.

\bibitem[Yang et~al.(2023)Yang, Ding, Lv, Jiang, He, Guo, Bai, and
  Tang]{yang2023gpt}
Zhen Yang, Ming Ding, Qingsong Lv, Zhihuan Jiang, Zehai He, Yuyi Guo, Jinfeng
  Bai, and Jie Tang.
\newblock Gpt can solve mathematical problems without a calculator.
\newblock \emph{arXiv preprint arXiv:2309.03241}, 2023.

\bibitem[Yao et~al.(2022)Yao, Zhao, Yu, Du, Shafran, Narasimhan, and
  Cao]{yao2022react}
Shunyu Yao, Jeffrey Zhao, Dian Yu, Nan Du, Izhak Shafran, Karthik~R Narasimhan,
  and Yuan Cao.
\newblock React: Synergizing reasoning and acting in language models.
\newblock In \emph{The Eleventh International Conference on Learning
  Representations}, 2022.

\bibitem[Yu et~al.(2023{\natexlab{a}})Yu, Gao, and Wang]{yu2023outcome}
Fei Yu, Anningzhe Gao, and Benyou Wang.
\newblock Outcome-supervised verifiers for planning in mathematical reasoning.
\newblock \emph{arXiv preprint arXiv:2311.09724}, 2023{\natexlab{a}}.

\bibitem[Yu et~al.(2023{\natexlab{b}})Yu, Jiang, Shi, Yu, Liu, Zhang, Kwok, Li,
  Weller, and Liu]{yu2023metamath}
Longhui Yu, Weisen Jiang, Han Shi, Jincheng Yu, Zhengying Liu, Yu~Zhang,
  James~T Kwok, Zhenguo Li, Adrian Weller, and Weiyang Liu.
\newblock Metamath: Bootstrap your own mathematical questions for large
  language models.
\newblock \emph{arXiv preprint arXiv:2309.12284}, 2023{\natexlab{b}}.

\bibitem[Yuan et~al.(2023)Yuan, Yuan, Li, Dong, Lu, Tan, Zhou, and
  Zhou]{yuan2023scaling}
Zheng Yuan, Hongyi Yuan, Chengpeng Li, Guanting Dong, Keming Lu, Chuanqi Tan,
  Chang Zhou, and Jingren Zhou.
\newblock Scaling relationship on learning mathematical reasoning with large
  language models, 2023.

\bibitem[Yue et~al.(2023)Yue, Qu, Zhang, Fu, Huang, Sun, Su, and
  Chen]{yue2023mammoth}
Xiang Yue, Xingwei Qu, Ge~Zhang, Yao Fu, Wenhao Huang, Huan Sun, Yu~Su, and
  Wenhu Chen.
\newblock Mammoth: Building math generalist models through hybrid instruction
  tuning.
\newblock \emph{arXiv preprint arXiv:2309.05653}, 2023.

\bibitem[Zhou et~al.(2023)Zhou, Wang, Lu, Shi, Luo, Qin, Lu, Jia, Song, Zhan,
  et~al.]{zhou2023solving}
Aojun Zhou, Ke~Wang, Zimu Lu, Weikang Shi, Sichun Luo, Zipeng Qin, Shaoqing Lu,
  Anya Jia, Linqi Song, Mingjie Zhan, et~al.
\newblock Solving challenging math word problems using gpt-4 code interpreter
  with code-based self-verification.
\newblock \emph{arXiv preprint arXiv:2308.07921}, 2023.

\end{thebibliography}
\bibliographystyle{plainnat}


\newpage
\section{Appendix}

\subsection{Introduction to the MARIO repository}

\subsubsection{Reproducing Data Generation}
\label{app:data}

To generate the solution of a provided question, please refer to the following example script in our submitted code repository.

\begin{verbatim}
python gpt_react.py \
--verbose \
--dataset math \
-g gpt-4-1106-preview \
-q "Find all the roots of x^4 + 4 = 0."    
\end{verbatim}

\subsubsection{Fine-tuning}

Our training is mostly performed on LLaMA-Factory\footnote{\url{https://github.com/hiyouga/LLaMA-Factory/)}} code base. Please refer to that repository for more details.

\subsubsection{Inference}

Single question inference.
\begin{verbatim}
python react.py -c /path/to/checkpoint_dir -q "Compute tan(45)." --verbose
\end{verbatim}

\noindent Large scale inference with vllm\footnote{\url{https://github.com/vllm-project/vllm}}.
\begin{verbatim}
python batch_react.py -c /path/to/checkpoint_dir -q /path/to/question_file
\end{verbatim}
Question file should be in \texttt{jsonl} format, where each line is a json string. The json string should at least include a key value pair for question.

\subsubsection{Evaluation Toolkit}

In order to evaluate the model prediction, it requires our implemented toolkit that is located in folder \texttt{math\_evaluation}.
\begin{verbatim}
python eval.py -q /path/to/question_file
\end{verbatim}
Question file should be in \texttt{jsonl} format, where each line is a json string at least containing "pred" and "answer" keys for prediction and ground truth, respectively.

\subsubsection{Prompts}
\label{app:prompts}

The REACT Instruction is as follows. For the demonstration examples for GSM8K and MATH, please refer to the file \texttt{prompts.py} in our submitted code repository.

\begin{verbatim}
You are a powerful agent with broad math knowledge and great Python programming 
skills. Answer the math questions as best you can. You have access to the following 
tool:

python_interpreter: A Python shell to execute python code snippet.

When solving math problem, you should think step by step, where each step includes 4 
mini-steps Thought/Action/Action Input/Observation. Note that if some step requires 
accurate calculation (including but not limited to symbolic simplification, 
derivation, numerical calculation, solving equations or inequalities), you should 
write Python code and execute it to obtain result. 
The following is the required template.

Question: the input question

Thought: the text analysis, and list the math equations if necessary

Action: the action to take, should be python_interpreter, or None

Action Input: the Python Code in markdown format (if Action is None, it is None), e.g.,
```python
import math
theta = math.pi / 4
some_var = math.cos(theta)
print(some_var)
```

Observation: the result of the action

... (this Thought/Action/Action Input/Observation can repeat N times)

Thought: the final analysis of the answer

Final Answer: the concise answer without verbose context

The following are 2 demonstrations examples.

{examples}

Now! It's your turn.

Question: {question}

Thought:
\end{verbatim}

\subsection{Human Review for MATH}
\label{app:human}

Our own researchers are responsible for the human review. 
We mainly fix the following issues in the field of final answer.

\begin{itemize}
    \item \textbf{Redundant text} Even in the prompt we have indicated the field of final answer should only include the math expression of final answer without other text. It is inevitable that a full sentence will be generated in this field. Therefore, we will remove the redundant text, \emph{e.g.},
    \begin{verbatim}
    Final Answer: John spent 25 dollars in total. => Final Answer: 25
    \end{verbatim}
    \item \textbf{Equivalent expression} Because the text analysis of LLM is based on python code snippets and the corresponding execution results, the generated final answers prefer the `sympy' format, which differs from the `latex' format provided in the MATH dataset. However, they are sometimes equivalent. In this case, we should consider the generated solution as correct, \emph{e.g.},
    \begin{verbatim}
    \\frac{8 - 7x}{6} = 4/3 - 7x/6
    \\begin{pmatrix} 1 & 2 \\\\ 3 & 4 \\end{pmatrix} = Matrix([[1, 2], [3, 4]])
    \end{verbatim}
    This also motivates us to develop the math evaluation toolkit. 
\end{itemize}

\newpage
\subsection{Re-formatting}
\label{app:reformat}

\begin{figure}[ht]
    \centering
    \includegraphics[width=\textwidth]{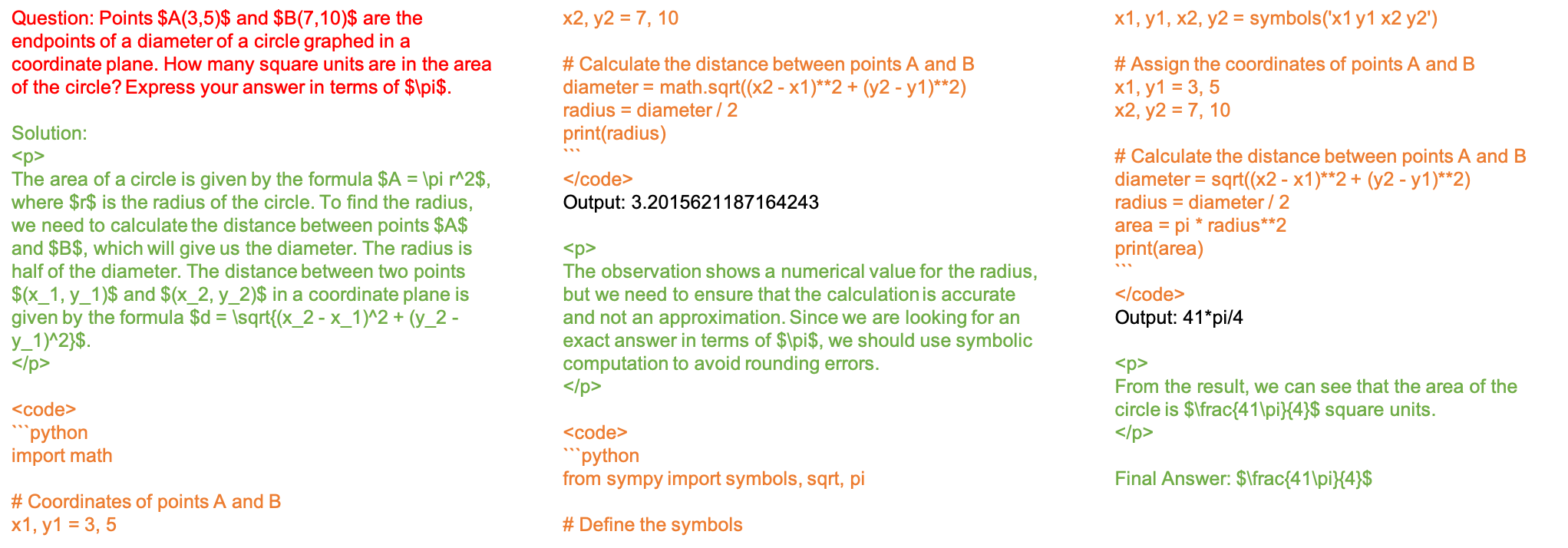}
    \caption{
    A re-formatting example includes a combined display of text analyses and accompanying code snippets. 
    Please note that the display represents a restructured format resembling HTML, from the original REACT format generated.
    }
    \label{fig:reformat}
\end{figure}

\subsection{Case Study on GSM-Hard}
\label{app:gsm_hard}

In this case study, we present three typical examples demonstrating why many of our model's final answers are unjustifiably deemed incorrect.

\subsubsection{Incorrect or unreasonable answer}

In this example, the PoT solution provided by GSM-Hard remains the same as the initial answer in the GSM8K test set, yet it fails to align with the question as it does not incorporate the number 780786 in its program. 
Meanwhile, the ToRA solution neglects the crucial detail that the quantity of Pokemon must be an integer. 
In contrast, our model had previously generated the fractional answer, but its internal self-verification process detected inconsistencies, leading it to ultimately output 'None.'

\begin{verbatim}
Question: Alex, Stan, and Adelwolfe are trying to catch them all, Pokemon that is. 
Together they have caught 780786 Pokemon. Alex has caught 5 more than Stan, and Stan 
has caught 13 less than 4 times as many as Adelwolfe has caught. How many Pokemon 
has Stan caught?

PoT Solution from GSM-Hard:

def solution():
    stan = -1  # we don't know how many pokemon stan has caught, so we set it to -1
    total = 780786
    alex = stan + 5
    adelwolfe = (stan + 13) / 4
    stan = total - alex - adelwolfe
    # write values for alex and adelwolfe in terms of stan
    stan = total - (stan + 5) - ((stan + 13) / 4)
    # simplify
    stan + 5 + (stan + 13) / 4 + stan == 339
    4 * stan + 20 + stan + 13 + 4 * stan == 4 * total
    9 * stan + 33 == 4 * total
    9 * stan == 4 * 339 - 33
    # so we get
    stan = (4 * 339 - 33) / 9
    result = stan
    return result
Answer: 147.0


PoT Solution from ToRA:

```python
from sympy import symbols, Eq, solve

def stan_pokemon():
    # Define the symbols
    a, s, ad = symbols('a s ad')
    
    # Define the equations
    eq1 = Eq(a + s + ad, 780786)
    eq2 = Eq(a, s + 5)
    eq3 = Eq(s, 4*ad - 13)
    
    # Solve the system of equations
    solution = solve((eq1, eq2, eq3), (a, s, ad))
    
    return solution[s]

result = stan_pokemon()
print(result)
```
```output
1041037/3
```
Stan has caught $\\boxed{\\frac{1041037}{3}}$ Pokemon.


Our Solution:

<p>
Let's assume that Adelwolfe has caught x Pokemon. According to the given information, Stan 
has caught 13 less than 4 times as many Pokemon as Adelwolfe, so Stan has caught (4x - 13) 
Pokemon. Alex has caught 5 more than Stan, so Alex has caught (4x - 13) + 5 = 4x - 8 Pokemon. 
Together, they have caught 780786 Pokemon, so we can set up the equation:

x + (4x - 13) + (4x - 8) = 780786

Now, I will solve this equation to find the value of x, which represents the number of 
Pokemon Adelwolfe has caught. Once I have x, I can calculate the number of Pokemon Stan 
has caught by substituting x into the expression (4x - 13).
</p>

<code>
```python
from sympy import symbols, Eq, solve

x = symbols('x')
equation = Eq(x + (4*x - 13) + (4*x - 8), 780786)
solution = solve(equation, x)
stan_pokemon = 4*solution[0] - 13
stan_pokemon
```
</code>
Output: 1041037/3

<p>
The solution to the equation is x = 1041037/3. However, since x represents the number of 
Pokemon Adelwolfe has caught, it must be an integer. I need to verify that the solution 
for x is indeed an integer.
</p>

<code>
```python
is_integer = solution[0].is_integer
is_integer
```
</code>
Output: False

Final Answer: None
\end{verbatim}

\subsubsection{Incorrect or unreasonable intermediate step}

In this case, the PoT solution offered by GSM-Hard calculates an intermediate variable cows as -3016191, which is not feasible. 
Even though the final answer representing the number of chickens is positive, the question is deemed invalid because the quantity of cows cannot be a negative value. 
Regarding the ToRA solution, it arrives at the same answer as GSM-Hard despite generating an incorrect program output. 
Our model's program produces identical intermediate results to what is considered the correct answer, yet upon textual analysis, it recognizes that the count of animals must not be negative. 
Consequently, the model persists in self-verification up to a predetermined maximum number of steps and ultimately yields 'None' as the output.

\begin{verbatim}
Question: Farmer Brown has 3016226 animals on his farm, all either chickens or cows. 
They have a total of 70 legs, all together. How many of the animals are chickens?

PoT Solution from GSM-Hard:

def solution():
    total_animals = 3016226
    total_legs = 70
    chicken_legs = 2
    cow_legs = 4
    cows = (total_legs - (total_animals * chicken_legs)) / (cow_legs - chicken_legs)
    chickens = total_animals - cows
    result = chickens
    return result
Answer: 6032417.0


PoT Solution from ToRA:

```python
def chicken_count():
    total_animals = 3016226
    total_legs = 70
    # Let\'s denote the number of chickens as x. Then the number of cows is 3016226 - x.
    # The total number of legs is the sum of the legs of the chickens and the cows.
    # So we have the equation: 2x + 4*3016226 = 70
    # Solving this equation for x gives us the number of chickens.
    chickens = (total_legs - 4 * total_animals) / 2
    return int(chickens)
    
chickens = chicken_count()
print(chickens)
```
```output
-6032417
```
The number of chickens is $\\boxed{6032417}$.


Our solution:

<p>
To solve this problem, we need to set up an equation based on the information given. 
We know that each chicken has 2 legs and each cow has 4 legs. Let's denote the number 
of chickens as $c$ and the number of cows as $w$. The total number of legs is the sum 
of the legs of all the chickens and all the cows, which can be expressed as $2c + 4w = 70$. 
We also know that the total number of animals is $c + w = 3016226$. We can use these 
two equations to solve for $c$ and $w$.
</p>

<code>
```python
from sympy import symbols, Eq, solve

# Define the symbols
c, w = symbols('c w')

# Equation for the total number of legs
legs_eq = Eq(2*c + 4*w, 70)

# Equation for the total number of animals
animals_eq = Eq(c + w, 3016226)

# Solve the system of equations
solution = solve((legs_eq, animals_eq), (c, w))
print(solution)
```
</code>
Output: {c: 6032417, w: -3016191}

<p>
The solution to the system of equations indicates that there are 6032417 cows and 
-3016191 chickens, which is not possible since the number of animals cannot be negative. 
This suggests there might be an error in the interpretation of the problem or in the 
equations set up. The negative number of chickens indicates that the equation for 
the total number of legs is incorrect. 
</p>

...(skip many verification steps)

Final Answer: None
\end{verbatim}

\subsubsection{Ambiguous question}

In this instance, the solutions provided by GSM-Hard and ToRA are more precise, yielding the exact figure of 377712.375. 
In fact, our model also generates this precise number during an intermediate step. 
However, the question's phrasing regarding ``how many hours'' is somewhat ambiguous, leading our model to round the answer to the nearest whole number to better align with the question's apparent emphasis on the duration in hours. 

\begin{verbatim}
Question: Lori wants to buy a $320.00 pair of shoes and a matching belt that is 
$3021379. Her part-time job pays her $8.00 an hour. How many hours will she have 
to work before she can make her purchase?

PoT Solution from GSM-Hard:

def solution():
    shoes_cost = 320
    belt_cost = 3021379
    hourly_wage = 8
    total_cost = shoes_cost + belt_cost
    hours_worked = total_cost / hourly_wage
    result = hours_worked    
    return result
Answer: 377712.375


PoT Solution from ToRA:

```python
def hours_to_work():
    shoe_cost = 320.00
    belt_cost = 3021379.00
    total_cost = shoe_cost + belt_cost
    hourly_wage = 8.00
    hours_needed = total_cost / hourly_wage
    return hours_needed
    
hours_needed = hours_to_work()
print(hours_needed)
```
```output
377712.375
```
Lori will have to work $\\boxed{377712.375}$ hours before she can make her purchase.


Our Solutions: 

<p>
To find out how many hours Lori needs to work, we need to calculate the total cost 
of the shoes and the belt, and then divide that by her hourly wage.
</p>

<code>
```python
shoes_cost = 320.00
belt_cost = 3021379
total_cost = shoes_cost + belt_cost
hourly_wage = 8.00
hours_needed = total_cost / hourly_wage\nprint(hours_needed)
```
</code>
Output: 377712.375

<p>
The calculation shows that Lori needs to work approximately 377712.375 hours to make 
her purchase. Since she can't work a fraction of an hour, she will need to round up 
to the nearest whole hour.
</p>

<p>
Since Lori can't work a fraction of an hour, and the calculation shows that she needs 
to work approximately 377712.375 hours, she will have to work 377713 hours to make 
her purchase.
</p>

Final Answer: $377713$
\end{verbatim}

\end{document}